\documentclass[11pt]{article}

\usepackage[preprint]{acl}

\usepackage{times}
\usepackage{latexsym}
\usepackage{verbatim}
\usepackage{amsmath}
\usepackage{amssymb}
\usepackage{booktabs}
\usepackage{hyperref}
\usepackage{multirow}
\usepackage[utf8]{inputenc}
\usepackage{graphicx}
\usepackage{booktabs}
\usepackage{tabularx}
\usepackage{xcolor}
\usepackage{subcaption}
\usepackage{listings}
\usepackage{amsmath}
\definecolor{codeblue}{rgb}{0.1,0.1,0.8}
\usepackage[T1]{fontenc}

\usepackage[utf8]{inputenc}

\usepackage{microtype}
\usepackage{xcolor}

\usepackage{inconsolata}

\usepackage{graphicx}

\title{TrackTeller: Temporal Multimodal 3D Grounding for Behavior-Dependent Object References}

\author{
  Jiahong Yu\textsuperscript{1}\thanks{These authors contributed equally to this work.}, 
  Ziqi Wang\textsuperscript{1}\footnotemark[1], 
  Hailiang Zhao\textsuperscript{1}\thanks{Corresponding author: hliangzhao@zju.edu.cn}, 
  Wei Zhai\textsuperscript{2}, 
  Xueqiang Yan\textsuperscript{3}, 
  Shuiguang Deng\textsuperscript{1} \\
  \normalsize
  \textsuperscript{1}Zhejiang University \\
  \textsuperscript{2}Fudan University \\
  \textsuperscript{3}Huawei Technologies Ltd. \\
}

\newcommand{\reviseyjh}[1]{\textcolor{red}{#1}}
\begin{document}
\maketitle
\begin{abstract}
Understanding natural-language references to objects in dynamic 3D driving scenes is essential for interactive autonomous systems. In practice, many referring expressions describe targets through recent motion or short-term interactions, which cannot be resolved from static appearance or geometry alone. We study temporal language-based 3D grounding, where the objective is to identify the referred object in the current frame by leveraging multi-frame observations. We propose TrackTeller, a temporal multimodal grounding framework that integrates LiDAR-image fusion, language-conditioned decoding, and temporal reasoning in a unified architecture. TrackTeller constructs a shared UniScene representation aligned with textual semantics, generates language-aware 3D proposals, and refines grounding decisions using motion history and short-term dynamics. Experiments on the NuPrompt benchmark demonstrate that TrackTeller consistently improves language-grounded tracking performance, outperforming strong baselines with a 70\% relative improvement in Average Multi-Object Tracking Accuracy and a 3.15–3.4× reduction in False Alarm Frequency.

\end{abstract}

\section{Introduction}
In autonomous driving and embodied perception tasks, linking natural-language expressions to objects in a 3D scene is a fundamental capability. While the task of 3D grounding has seen considerable progress in indoor and static settings, it remains under-explored for dynamic, real-world driving scenarios. Prior work such as LanguageRefer \citep{roh2022languagerefer} demonstrates how spatial-language models can identify objects in point clouds from a single static scene. NS3D  \citep{hsu2023ns3d} extend grounding to multi-object relations with neuro-symbolic reasoning. Multi-View Transformer \citep{huang2022multi} enhances robustness via multiple views of static 3D scenes. Meanwhile, large-scale driving-domain datasets such as Talk2Car \citep{deruyttere2022talk2car} built on nuScenes\footnote{\url{https://www.nuscenes.org/}} have leveraged natural language in multimodal driving contexts. However, these works focus primarily on single-frame snapshots or static references and do not address the problem of grounding expressions that refer to objects by their recent motion, behavior, or temporal interactions.

\begin{figure}
    \centering
    \includegraphics[width=0.94\linewidth]{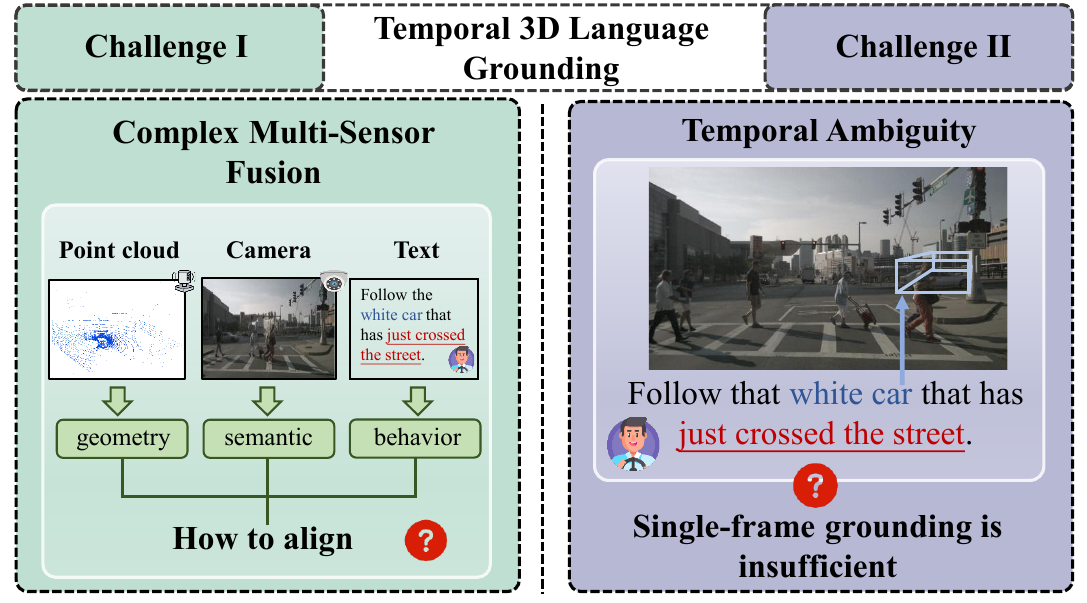}
    \caption{Key challenges of temporal 3D language grounding. (I) Complex multi-sensor fusion: grounding requires coherent alignment of LiDAR geometry, camera semantics, and language cues. (II) Temporal ambiguity: many expressions reference recent motion or behavior, requiring multi-frame reasoning.}
    \label{fig:intro}
\end{figure}

Addressing this gap requires meeting two key challenges shown in Fig. \ref{fig:intro}. First, autonomous driving requires multimodal perception that combines LiDAR geometry, camera semantics, along with efficient fusion mechanisms that align geometric cues, visual semantics, and language descriptions. BEV representations have proven effective for detection and tracking \citep{feng2025lst} but have rarely been integrated with language grounding. Second, language often refers to an object in terms of its temporal behavior (e.g., ``Follow that white car that just crossed the street.''), which cannot be resolved by static frame analysis alone. Existing 3D grounding models, including recently sequential grounding modules like GroundFlow \citep{lin2025groundflow} underscore the importance of modeling history but remain outside the driving scenario. In addition, these approaches are designed for point-cloud sequences in controlled environments and do not extend to the multi-sensor, motion-rich nature of driving scenes. 

In this work we propose TrackTeller, a temporal multimodal grounding framework designed for autonomous driving scenes. TrackTeller integrates LiDAR and multi-view images into a unified UniScene representation, aligns language semantics with this multimodal token space, and generates 3D candidate proposals via a language-conditioned decoder. To resolve expressions that depend on motion or recent interactions, a temporal reasoning module enriches proposal embeddings with historical context and short-term dynamics.

Our contributions are threefold: (1) We formulate temporal language-based 3D grounding for autonomous driving, where objects are grounded in the current frame by leveraging multi-frame observations to interpret motion- and behavior-dependent expressions. (2) We propose a language-aligned 3D decoding framework that injects textual semantics into a unified UniScene representation and enables text-conditioned proposal generation via attention-based query refinement. (3) We introduce a language-aware temporal reasoning module that incorporates memory attention and motion cues to robustly ground objects described by short-term dynamics and interactions. Experiments on the NuPrompt dataset demonstrate that TrackTeller effectively localizes objects described by dynamic expressions, 
significantly outperforming existing baselines and enabling real-time 
language-driven perception in autonomous driving.


\section{Related Work}
Language-guided visual grounding focuses on localizing objects referred to by natural-language expressions. Early research primarily addresses 2D images, where the task is to identify image regions described by text. As 3D perception matures, grounding extends to point clouds and reconstructed 3D scenes, giving rise to text-guided 3D visual grounding. A recent survey by Liu et al.~\citep{liu2025survey} provides a comprehensive overview of this field, summarizing task formulations, multimodal alignment strategies, and the challenges of complex spatial-semantic reasoning. ViewRefer~\citep{guo2023viewrefer} explores how multi-view visual cues can be combined with language to improve the robustness of 3D object localization.

In the context of autonomous driving, grounding evolves toward large-scale environments. NuGrounding~\citep{li2025nugrounding} introduces a framework that integrates multi-camera imagery with 3D geometry for language-guided grounding in driving scenes. SeeGround~\citep{li2025seeground} enables zero-shot open-vocabulary 3D grounding by leveraging pretrained 2D vision-language models together with differentiable rendering. Beyond grounding itself, language is also studied as a high-level interface for perception in driving systems. OmniDrive~\citep{wang2024omnidrive} presents an LLM-based agent that unifies 3D perception, reasoning, and planning, while language-prompted driving~\citep{wu2025language} and embodied driving understanding~\citep{zhou2024embodied} investigate how linguistic inputs enhance scene understanding and downstream driving behaviors.

Despite these advances, most existing 3D grounding approaches operate on single frames or static scene representations. However, many expressions in driving scenarios inherently refer to temporal behaviors. GroundFlow~\citep{lin2025groundflow} introduces temporal reasoning for sequential point-cloud grounding and demonstrates the importance of modeling history, but its design targets controlled point-cloud sequences rather than multimodal autonomous driving pipelines. Fine-grained grounding methods~\citep{dey2025fine} incorporate more detailed spatial and linguistic constraints, yet they still lack explicit reasoning over object motion.

Existing studies highlight the importance of multimodality, spatial reasoning, and language alignment for 3D grounding. However, temporal language-based grounding in real-world driving environments remains largely underexplored. Resolving expressions that depend on recent motion or behavior requires joint modeling of cross-frame dynamics, multimodal scene representations, and language-conditioned reasoning. This work addresses this gap by proposing a unified framework for temporal multimodal 3D grounding in autonomous driving scenarios.

\section{Method}
\begin{figure*}
    \centering
    \includegraphics[width=1\linewidth]{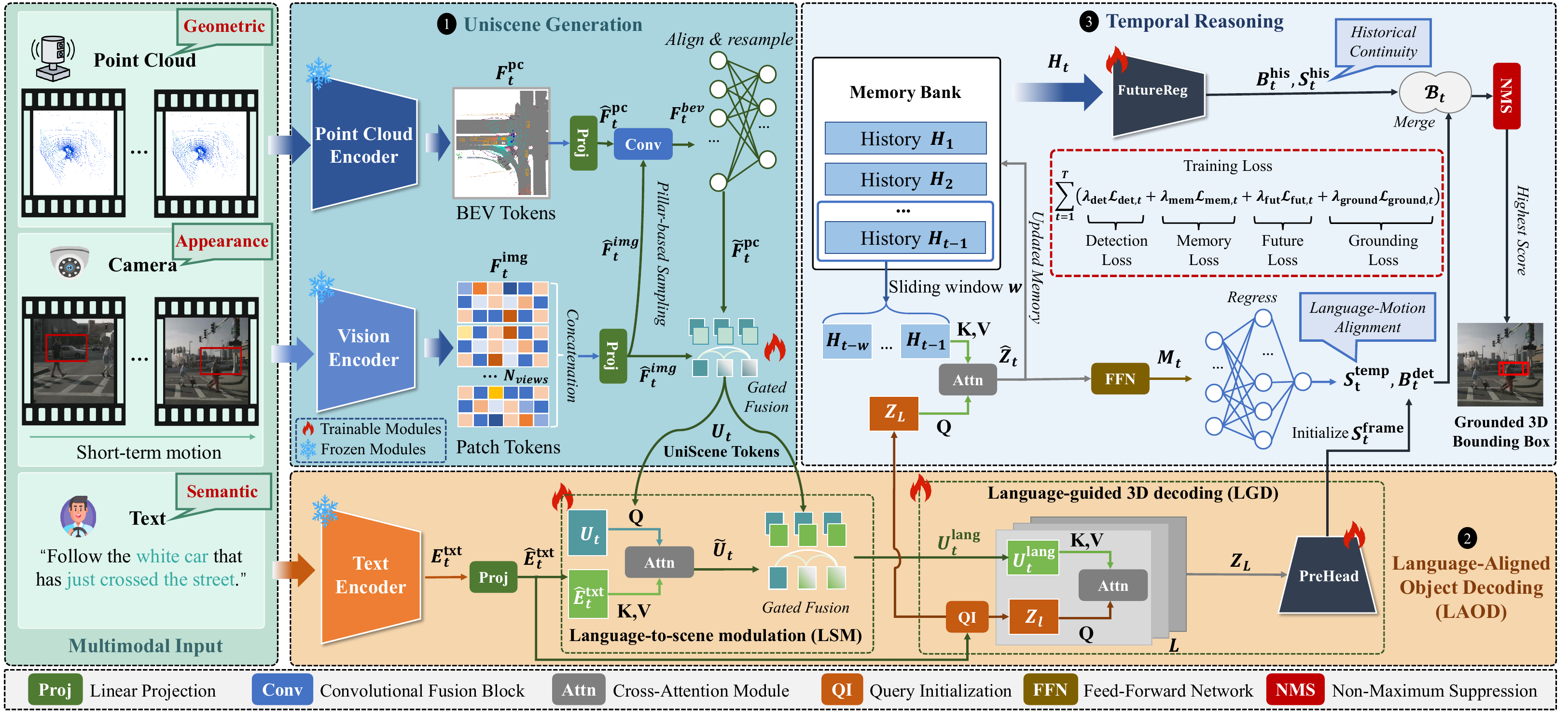}
    \caption{Overview of TrackTeller. LiDAR and multi-view images are fused into UniScene tokens, which are aligned with language for 3D proposal decoding. A temporal reasoning module enriches proposals with motion history and produces the final grounding score.}
    \label{fig:arch}
\end{figure*}

TrackTeller is a unified framework for temporal language-based 3D grounding. 
For each frame, LiDAR and multi-view images are fused into a set of UniScene Tokens, preserving both geometric and semantic information. The referring expression is aligned with these tokens, and a query-based decoder generates 3D bounding box proposals. A memory reasoning module further incorporates historical context to capture short-term motion and maintain object consistency across frames. The final grounded object is selected based on language-conditioned temporal proposal scores. The overall architecture is illustrated in Fig.~\ref{fig:arch}.

\subsection{UniScene Generation}
Given multi-sensor observations at time step $t$, our goal is to construct a unified scene representation that preserves geometric cues from LiDAR and semantic richness from multi-view images. Let $P_t$ denote the LiDAR point cloud and $I_t = \{ I_t^{(k)} \}_{k=1}^{N_{\text{views}}}$ denote the $N_{\text{views}}$ camera views at time $t$.

Following standard practice in LiDAR-camera multimodal 3D detection, the point cloud is first processed by a point cloud encoder \cite{yin2024fusion} to produce Bird’s Eye View (BEV) tokens: $F_t^{\text{pc}} \in \mathbb{R}^{N_{\text{bev}} \times D_{\text{pc}}}$, where $N_{\text{bev}}$ is the number of BEV tokens produced by the encoder, and $D_{\text{pc}}$ denotes the feature dimension of each token. In addition, each camera image is encoded into patch tokens by a pre-trained vision backbone \cite{pang2023standing}, and tokens from all views are concatenated as
\begin{equation}
F_t^{\text{img}} \in \mathbb{R}^{(N_{\text{views}} N_{\text{patches}}) \times D_{\text{img}}}.
\end{equation}
where $N_{\text{patches}}$ is the number of tokens extracted from each view by the vision encoder, and $D_{\text{img}}$ is the feature dimension of each patch token.

To obtain a geometry-aware yet semantically expressive representation, we project both modalities into a shared embedding space with dimension $D$ and adopt a gated fusion strategy. Specifically,
\begin{equation}
\hat{F}_t^{\text{pc}} = W_{\text{pc}} F_t^{\text{pc}}, \qquad
\hat{F}_t^{\text{img}} = W_{\text{img}} F_t^{\text{img}},
\end{equation}
where $W_{\text{pc}}$ and $W_{\text{img}}$ are learnable projections. Image features are then lifted into the BEV space via a pillar-based sampling ($F_t^{\mathrm{img}\to\mathrm{bev}}$) and fused with BEV tokens through a convolutional block:
\begin{equation}
F_t^{\mathrm{bev}} =
\mathrm{Conv}\!\left(
[\, F_t^{\mathrm{img}\to\mathrm{bev}} \Vert \hat{F}_t^{\mathrm{pc}} \,]
\right)\in \mathbb{R}^{N_{\mathrm{bev}} \times D}.
\end{equation}
where $[\cdot\Vert \cdot]$ denotes channel-wise concatenation. The fused BEV representation is subsequently aligned back to the image token resolution by assuming a ground-plane geometry where each image pixel $(u, v)$ is back-projected to a 3D point $(x, y, 0)$ and subsequently mapped to its corresponding BEV coordinate. We then resample the BEV grid to match the image resolution:
\begin{equation}
\tilde{F}_t^{\mathrm{pc}} =
\mathrm{Align}_{\mathrm{pc}}\!\left(F_t^{\mathrm{bev}}\right) \in \mathbb{R}^{(N_{\mathrm{views}} N_{\mathrm{patches}}) \times D}
\end{equation}

A gating module adaptively balances geometric and semantic contributions:
\begin{equation}
\label{gate}
G_t = \sigma\!\left( W_g [\, \tilde{F}_t^{\text{pc}} \Vert \hat{F}_t^{\text{img}} \,] \right),
\end{equation}
where $\sigma(\cdot)$ denotes the sigmoid activation and $W_g$ is a learnable projection. The final fused representation is obtained by:
\begin{equation}
U_t = G_t \odot \tilde{F}_t^{\text{pc}} + (1 - G_t) \odot \hat{F}_t^{\text{img}}.
\end{equation}

We refer to $U_t$ as UniScene Tokens, a unified multimodal token representation that provides geometry-aware scene embeddings. 

\subsection{Language-Aligned Object Decoding}
The goal of this module is to integrate linguistic semantics into UniScene tokens and to generate language-conditioned 3D bounding box proposals. TrackTeller performs language alignment and object decoding in a unified process: (1) linguistic modulation of UniScene Tokens, followed by (2) language-guided decoding via a PETR-based head~\cite{liu2022petr} that produces 3D bounding box proposals and grounding scores. 

\subsubsection{Language-to-scene modulation (LSM)}

Given a natural-language referring expression 
$q = (w_1,\dots,w_{L_{txt}})$, where $w_i$ denotes the $i$-th token of the input sentence
(e.g., ``Follow the white car that just crossed the intersection''), a text encoder \cite{liu2019roberta} produces token embeddings
\begin{equation}
E_t^{\text{txt}} = \mathrm{TextEnc}(q) \in \mathbb{R}^{L_{txt} \times D_{\text{txt}}},
\end{equation}
which are then projected into the shared token space with dimension $D$: $\hat{E}_t^{\text{txt}} = W_{\text{txt}} E_t^{\text{txt}}$. To inject linguistic information, a gated cross-attention module is applied. First, a standard cross-attention ($\mathrm{CA}$) retrieves textual features using $U_t$ as queries:
\begin{equation}
\tilde{U}_t = \mathrm{CA}\bigl(Q = U_t , K = \hat{E}_t^{\text{txt}} , V = \hat{E}_t^{\text{txt}}\bigr).
\end{equation}
The final language-aware embedding $U_t^{\text{lang}}$ is obtained by fusing $U_t$ and $\tilde{U}_t$ through the same gated fusion mechanism as described in (\ref{gate}), where linguistic cues act as adaptive refinements to the scene representation.

\subsubsection{Language-guided 3D decoding (LGD)}
The detection head operates on $U_t^{\text{lang}}$ to generate 3D bounding box proposals and their associated embeddings. We initialize a set of object queries
$Z_0 \in \mathbb{R}^{M \times D}$ from a learned query embedding matrix $Z_{\text{init}}$, where $M$ denotes the number of queries. To inject sentence-level semantics, text embeddings
$\hat{E}_t^{\text{txt}}$ are first aggregated by mean pooling into
$e_t^{\text{txt}} \in \mathbb{R}^{D}$.
A linear projection $W_q \in \mathbb{R}^{D \times D}$ maps this vector into the query space and conditions all queries as $Z_0 = Z_{\text{init}} + W_q\, e_t^{\text{txt}}.$ This conditioning biases the queries toward objects consistent with the referring expression.

The decoder then iteratively refines the queries through cross-attention over the language-aware scene representation.
Let $Z_l$ denote the queries at the $l$-th decoder layer, where each layer performs
\begin{equation}
Z_{l+1} = \mathrm{CA}\bigl(
Q {=} Z_l,\;
K {=} U_t^{\text{lang}},\;
V {=} U_t^{\text{lang}}
\bigr).
\end{equation}
Through this process, the queries progressively integrate geometric evidence and linguistic cues. The final decoder outputs $Z_L$ are fed into a prediction head $\mathrm{PreHead}(\cdot)$ to regress 3D bounding boxes and produce frame-level grounding scores:
\begin{equation}
B_t^{\text{det}},\; S_t^{\text{frame}} = \mathrm{PreHead}(Z_L),
\end{equation}
where $B_t^{\text{det}}$ denotes the set of candidate 3D boxes and $S_t^{\text{frame}}$ provides the initial language-conditioned scores, which are further refined by the temporal reasoning module. The detail of $\mathrm{PreHead}(\cdot)$ is illustrated in the Appendix \ref{app:prehead}.

\subsection{Temporal Reasoning}
While the LGD provides per-frame grounding scores, many referring expressions in driving scenarios depend on short-term temporal cues
(e.g., motion patterns or recent interactions) that cannot be resolved from a
single frame. To incorporate these dependencies, TrackTeller includes a
temporal reasoning module \cite{pang2023standing} that enriches proposal embeddings with historical
context and evaluates whether their recent behavior matches the referring
expression \cite{wu2025language}.

\subsubsection{Historical Reasoning}
For each candidate proposal, the model maintains a memory bank that stores object-level embeddings from previous frames. Let $H_{t-1}$ denote this memory at time $t\!-\!1$. Given the current proposal embeddings $Z_L$, we update them by attending to $H_{t-1}$ to retrieve relevant historical context:
\begin{equation}
\hat{Z}_t =
\mathrm{CA}(Q {=} Z_L,\; K {=} H_{t-1},\; V {=} H_{t-1}),
\end{equation}
which injects motion history into the current embeddings. The memory is then updated by replacing the stored embeddings with $\hat{Z}_t$. The enriched embeddings are further refined by a lightweight temporal encoder: $M_t = \mathrm{TempEnc}(\hat{Z}_t)$, where $\mathrm{TempEnc}(\cdot)$ is a feedforward network that refines short-term motion patterns inferred from the retrieved memory.

\subsubsection{Temporal Grounding Prediction}
To maintain the continuity of object hypotheses, future reasoning is performed to propagate historical proposals across frames.
Given the historical embeddings $H_t$, a future prediction module propagates historical proposals to the current frame:
\begin{equation} 
B_{t}^{\text{his}},\, S_{t}^{\text{his}} = \mathrm{FutureReg}(H_t),
\end{equation}
where $\mathrm{FutureReg}(\cdot)$ predicts future proposals and their confidence scores based on past motion patterns. Specific implementation details are provided in the Appendix \ref{appendix:future_reasoning}.

In parallel, the motion-aware proposal embeddings $M_t$ obtained from historical reasoning are mapped to temporal grounding scores through a lightweight Multilayer Perceptron (MLP), obtaining $S_t^{\text{temp}}$. It evaluates how well each proposal’s recent motion aligns with the referring expression.

For each frame, the propagated historical 3D boxes $B_t^{\text{his}}$ with confidence scores $S_t^{\text{his}}$ and the current-frame $B_t^{\text{det}}$ with language- and motion-conditioned scores $S_t^{\text{temp}}$ are merged into a unified candidate set $\mathcal{B}_t$. Non-maximum suppression \cite{pang2025spatiotemporal} is applied to remove redundant hypotheses. The grounded object is selected as the proposal with the highest score, corresponding to the candidate whose recent motion dynamics and language-conditioned evidence best match the referring expression.

\subsection{Unified Multi-Task Learning Objective}

The total training loss is defined as a multi-task objective that jointly supervises object detection, temporal memory consistency, future proposal forecasting, and language-conditioned grounding across frames.
The overall loss is formulated as:
\begin{align}
    \mathcal{L}_{\text{total}} = \sum_{t=1}^{T} ( \lambda_{\text{det}} \mathcal{L}_{\text{det}, t} + \lambda_{\text{mem}}\mathcal{L}_{\text{mem}, t}  \notag \\
+ \lambda_{\text{fut}} \mathcal{L}_{\text{fut}, t} + \lambda_{\text{ground}} \mathcal{L}_{\text{ground}, t} ),
\end{align}
where each loss term corresponds to a specific component of TrackTeller:
\begin{itemize}
\item \textbf{Detection Loss ($\mathcal{L}_{\text{det}, t}$):}
It consists of a focal loss for object classification and an $L_1$ loss for 3D bounding box regression of candidate proposals $B_t^{\text{det}}$.

\item \textbf{Memory Loss ($\mathcal{L}_{\text{mem}, t}$):}
It enforces temporal consistency of $M_t$ after historical reasoning by encouraging alignment between temporally associated instances.

\item \textbf{Future Forecasting Loss ($\mathcal{L}_{\text{fut}, t}$):}
An $L_1$ loss is applied to supervise the predicted future 3D bounding boxes:
\begin{equation}
\mathcal{L}_{\text{fut}, t}
= \mathbf{A}_t \, \bigl\| B_{t}^{\text{his}} - B_{t}^{\text{gt}} \bigr\|_1,
\end{equation}
where $\mathbf{A}_t$ is a binary mask indicating active track instances. $B_t^{\text{gt}}$ denotes the ground-truth 3D bounding box at time step $t$.

\item \textbf{Grounding Loss ($\mathcal{L}_{\text{ground}, t}$):}
A focal loss is used to supervise the temporal grounding scores $S_t^{\text{temp}}$, aligning the highest-scoring proposal with the ground-truth object referred to in the expression.
\end{itemize}

\section{Experimental Results and Discussion}
This section outlines the experimental protocol and presents comparative results for TrackTeller.

\subsection{Experimental Setup}
\subsubsection{Dataset and Task}
\begin{figure}
    \centering
    \includegraphics[width=0.95\linewidth]{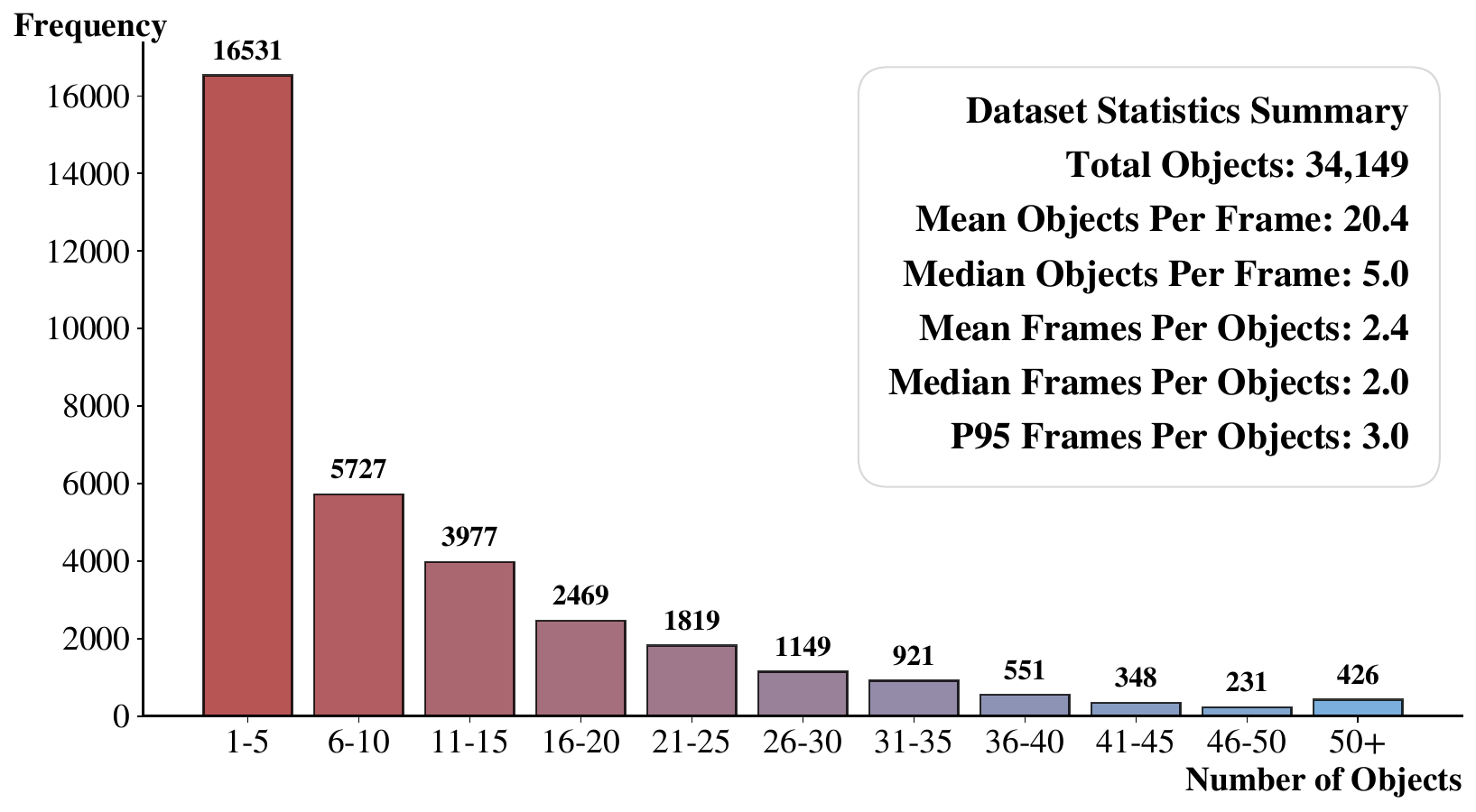}
    \caption{Number of referred targets per prompt.}
    \label{fig:bucket}
\end{figure}

\begin{figure}
    \centering
    \includegraphics[width=0.95\linewidth]{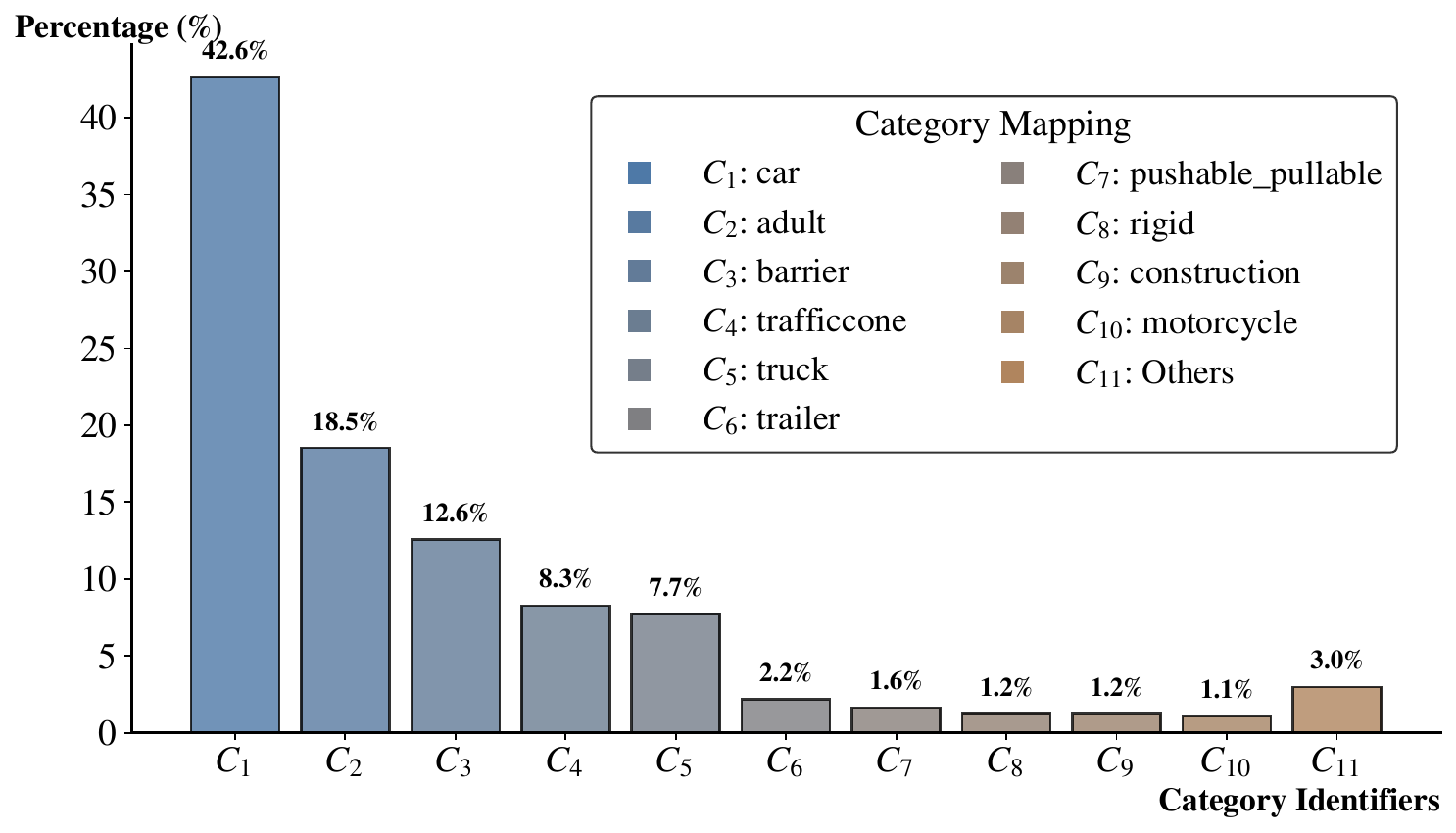}
    \caption{Percentage of referred targets per category.}
    \label{fig:category}
\end{figure}

We adopt the NuPrompt dataset~\cite{wu2025language}, which is built upon nuScenes and provides natural-language referring expressions for dynamic traffic scenes. NuPrompt consists of 850 scenes, 34,149 frames, and 40,776 prompts, spanning diverse cities, weather conditions, and illumination settings. Compared to the original image-text-only formulation, we additionally incorporate synchronized LiDAR point clouds for each frame, yielding a more challenging multimodal and temporal grounding task. 
We analyze the dataset from two complementary perspectives shown in Figs.~\ref{fig:bucket} and~\ref{fig:category}. 


Fig.~\ref{fig:bucket} shows the distribution of the number of referred targets per prompt. While most prompts involve few objects, a non-negligible portion refers to dozens of instances, indicating frequent multi-instance and potentially ambiguous grounding scenarios. Fig.~\ref{fig:category} presents the category distribution of referred targets. Vehicles, particularly cars, dominate the prompts and account for more than 40\% of all references, followed by adult pedestrians and infrastructure-related objects such as barriers and traffic cones. This distribution reflects both the prevalence of common traffic participants and the diversity of object types in real-world scenes. Data examples are provided in the Appendix \ref{dataex}.

\subsubsection{Implementation Details}
\label{sec:implementation details}
In our implementation, the UniScene representation has a spatial resolution of $20 \times 50$ with 256 channels. Both the LSM and LGD modules use a single-layer transformer with 8 attention heads and a hidden dimension of 2048. The prediction head is a six-layer PETRTransformer~\cite{liu2022petr}, with 8 attention heads and a 2048-dimensional feedforward network per layer. Temporal reasoning is implemented using separate single-layer transformer decoders, each with 8 heads and an embedding dimension of 256. The model is trained using Adam with an initial learning rate of $2\times10^{-4}$. The loss weights are set to $\lambda_{\text{det}}$ (cls: 2.0, bbox: 0.25), $\lambda_{\text{mem}}=2.0$, $\lambda_{\text{fut}}=0.5$, and $\lambda_{\text{ground}}=1.0$. All experiments are conducted for 15 epochs on a single NVIDIA RTX 4090 GPU. Additional results on parameter sensitivity are reported in Appendix~\ref{app:param_sensitivity}.


\begin{table*}[t]
\centering

\scriptsize
\setlength{\tabcolsep}{1pt}
\begin{tabular}{c|l|cccc|cccc|cccc|cccc|cccc}
\toprule
 &  & \multicolumn{20}{c}{\textbf{Filtering Threshold $\tau$}} \\
Modality & Method
& \multicolumn{4}{c}{AMOTA$\uparrow$}
& \multicolumn{4}{c}{Recall$\uparrow$}
& \multicolumn{4}{c}{AMOTP$\downarrow$}
& \multicolumn{4}{c}{TID$\downarrow$}
& \multicolumn{4}{c}{FAF$\downarrow$}\\

\cmidrule(lr){3-22}
 & 
 & 0 & 0.1 & 0.2 & 0.3
 & 0 & 0.1 & 0.2 & 0.3
 & 0 & 0.1 & 0.2 &0.3
 & 0 & 0.1 & 0.2 & 0.3
 & 0 & 0.1 & 0.2 &0.3\\

\midrule

\multicolumn{20}{c}{\textbf{DQTrack Variants}} \\
\midrule
Img & DQTrack (DETR3D)
& 0.43 & 0.51 & 0.87 &  0.26
& 33.66 & 23.65 & 8.67 & 3.07
& 1.65 & 1.78 & 1.93&  1.97
& 3.08 & 4.62 & 10.38 & 16.5
& 3368.9 & 1991.3 & 658.2 & 501.5 \\

Img & DQTrack (PETR)
& 0.24 & 0.28 & 0.31& 0.26
& 31.38 & 25.77 & 9.90 & 3.07
& 1.66 & 1.77 & 1.90 & 1.97
& 3.46 & 4.61 & 10.01 & 16.53
& 3376.3 & 2050.2 & 747.5 & 501.2\\

\midrule
\multicolumn{20}{c}{\textbf{PF-Track Variants}} \\
\midrule
Img & PF-Track (DETR3D)
& 0.08 & 0.06 & 1.73 & 11.05
& 50.57 & 38.53 & 17.29 & 20.99
& 1.38 & 1.54 & 1.78 & 1.75
& 2.31 & 3.46 & 8.85 & 8.85
& 8292.2 & 3035.8 & 764.7 & 253.9 \\

Img & PF-Track (PETR)
& 0.03 & 0.15 & 1.23 & 1.66
& 45.5 & 35.8 & 22.3 & 8.54
& 1.42 & 1.57 & 1.77 & 1.92
& 1.15 & 2.69 & 6.15 & 12.31
& 7194.2 & 3195.3 & 911.4  & 464.9\\

\midrule
\multicolumn{20}{c}{\textbf{PromptTracker}} \\
\midrule
Img & PromptTrack (DETR3D)
& 0.04 & 0.49 & 2.14 & 3.22
& 47.78 & 35.18 & 19.85 & 9.50
& 1.40 & 1.58 & 1.74 & 1.88
& 2.31 & 3.85 & 5.77 & 11.53
& 6467.7 & 2204.0 & 705.5  &  422.0\\

Img & PromptTrack (PETR)
& 6.97 & 10.62 & 1.10 & 1.67
& 43.55 & 42.26 & 17.43 & 8.54
& 1.41 & 1.42 & 1.81 & 1.92
& 3.08 & 3.08 & 7.69 & 12.31 
& 673.1 & 361.7 & 473.9 &  464.8\\

\midrule
\multicolumn{20}{c}{\textbf{PromptTracker-3D}} \\
\midrule
Img+LiD & PromptTrack-3D (DETR3D)
& 0.47 & 0.83 & 1.10 &  2.85
& 47.48 & 30.45 & 17.43 & 8.20
& 1.41 & 1.65 & 1.81 &  1.91
& 1.53 & 4.23 & 7.69 & 13.85
& 6134.6 & 1767.9 & 473.9 & 402.2 \\

Img+LiD & PromptTrack-3D (PETR)
& 7.08 & 7.37 & 7.08& 0.41
& 50.80 & 40.18 & 37.69 & 2.33
& 1.39 & 1.51 & 1.55 & 1.98
& 2.96 & 2.97 & 5.03 & 16.92
& 720.4 & 729.1 & 723.1 & 506.9\\

\midrule
\multicolumn{20}{c}{\textbf{GroundTeller (Ours)}} \\
\midrule
Img+LiD & \textbf{TrackTeller}
& \textbf{5.94} & \textbf{15.79} & \textbf{17.16}& \textbf{18.79}
& \textbf{48.21} & \textbf{44.34} & \textbf{41.24} & \textbf{28.42}
& \textbf{1.34} & \textbf{1.35} & \textbf{1.42} & \textbf{1.63}
& \textbf{2.69} & \textbf{2.69} & \textbf{3.85} & \textbf{8.08}
& \textbf{719.4} & \textbf{255.9} & \textbf{187.7}  & \textbf{147.6} \\

\bottomrule
\end{tabular}
\caption{Comprehensive comparison of prompt-guided tracking performance under different filtering thresholds $\tau$. ``Img'' denotes camera-only models, while ``Img+LiD'' represents multimodal approaches that jointly leverage image and point cloud inputs.}
\label{tab:full_metrics_mtml}
\end{table*}

\begin{figure*}[htbp!]
    \centering
    \includegraphics[width=\textwidth]{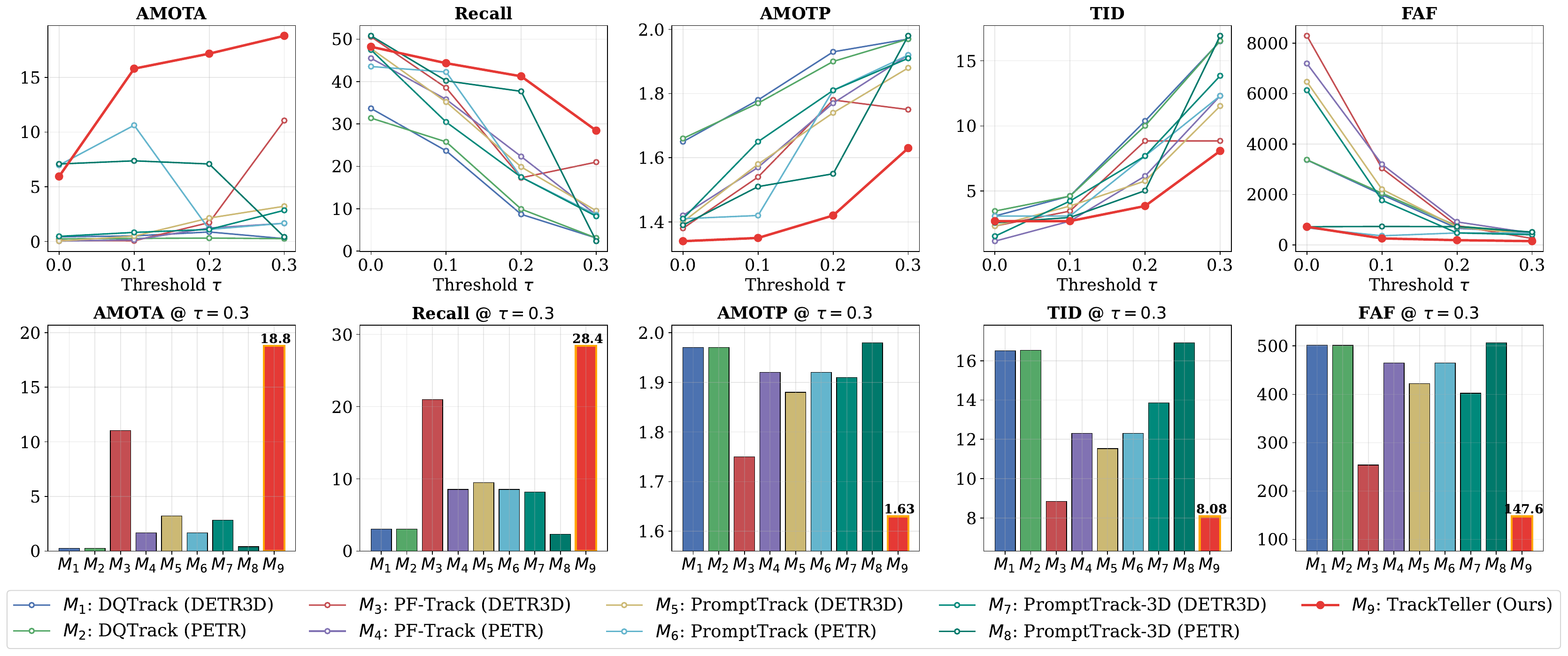}
    \caption{Performance comparison with representative baselines across filtering thresholds. As the filtering threshold increases, the number of effective candidate targets decreases sharply, which in turn leads to a notable degradation in tracking performance. In practice, threshold values larger than 0.3 substantially reduce overall accuracy. Therefore, we evaluate all methods under threshold values ranging from 0 to 0.3.}
    \label{fig:main results}
\end{figure*}

\subsection{Evaluation Metrics}
We evaluate TrackTeller using standard metrics for prompt-guided multi-object tracking. Our primary metric is AMOTA \cite{wang2024onetrack}, which summarizes tracking performance by jointly accounting for false negatives, false positives, and identity errors across recall thresholds, providing a robust evaluation of temporal tracking quality. We additionally report AMOTP \cite{sunderraman2024uav3d} to measure 3D localization accuracy, and Recall \cite{tan2025driver} to quantify the coverage of prompt-referred ground-truth objects.

To assess temporal consistency under language conditioning, we introduce Temporal Identity Discontinuity (TID) \cite{faggioli2024identity}, which measures trajectory fragmentation over time, where lower values indicate more stable identity preservation. We further report False Alarm Frequency (FAF) \cite{ward2024leveraging}, which reflects robustness to prompt-irrelevant distractors by measuring false positives per frame. All metrics are reported under different filtering thresholds to analyze the trade-off between coverage and tracking stability. Formal definitions of all metrics are provided in Appendix~\ref{EM}.



\subsection{Baselines}
We evaluate our approach against several query-based tracking baselines, i.e., 

\begin{itemize}
    \item \textbf{DQTrack}~\cite{li2023end} employs decoupled object queries to mitigate task conflicts in end-to-end query-based 3D tracking frameworks.
    \item \textbf{PFTrack}~\cite{pang2023standing} adopts a tracking-by-attention framework with object queries. The \textit{Past Reasoning} module updates object representations via cross-attention over historical queries, while the \textit{Future Reasoning} module predicts future object trajectories.
    \item \textbf{PromptTrack}~\cite{wu2025language} is the first method to leverage prompt-image fusion for tracking in nuScenes. A prompt reasoning branch is designed to predict the prompt-referred objects.
    \item \textbf{PromptTrack-3D}~\cite{wu2025language} is built upon PromptTrack~\cite{wu2025language} and IS-Fusion~\cite{yin2024fusion} by fusing LiDAR and image features for tracking.
\end{itemize}

The implementation details of baseline methods are provided in the Appendix~\ref{app:baseline_heads}.


\begin{table*}[t!]
\centering
\small
\setlength{\tabcolsep}{5pt} 
\begin{tabular}{cc|cc|cc|cc|ccccc}
\toprule
\multicolumn{2}{c|}{Backbone} & \multicolumn{2}{c|}{UG} & \multicolumn{2}{c|}{LAOD} & \multicolumn{2}{c|}{TR} & \multicolumn{5}{c}{Metrics} \\
Image & Prompt & Img & LiD & LSM & LGD & HR & TGP & AMOTA$\uparrow$ & Rec.$\uparrow$ & AMOTP$\downarrow$ & TID$\downarrow$ & FAF$\downarrow$ \\ \midrule
VoVNet & RoBERTa & \checkmark & & & & & & 4.20 & 10.38 & 1.92 & 12.75 & 367.9 \\ 
VoVNet & RoBERTa & \checkmark & \checkmark & & & & & 7.80 & 12.79 & 1.86 & 12.72 & 345.7 \\ 
VoVNet & RoBERTa & \checkmark & \checkmark & \checkmark & & & & 10.10 & 18.20 & 1.78 & 10.84 & 312.5 \\ 
VoVNet & RoBERTa & \checkmark & \checkmark & \checkmark & \checkmark & & & 12.10 & 21.61 & 1.72 & 9.67 & 281.4 \\ 
VoVNet & RoBERTa & \checkmark & \checkmark & \checkmark & \checkmark & \checkmark & & 15.23 & 25.46 & 1.71 & 8.98 & 256.2 \\ 
VoVNet & RoBERTa & \checkmark & \checkmark & \checkmark & \checkmark & \checkmark & \checkmark & \textbf{18.79} & \textbf{28.42} & \textbf{1.70} & \textbf{8.08} & \textbf{147.6} \\ \midrule
ResNet-101 & RoBERTa & \checkmark & \checkmark & \checkmark & \checkmark & \checkmark & \checkmark & 13.15 & 21.8 & 1.75 & 9.12 & 265.4 \\ 

VoVNet & CLIP & \checkmark & \checkmark & \checkmark & \checkmark & \checkmark & \checkmark & 15.42 & 25.2 & 1.74 & 8.59 & 221.3\\ \bottomrule
\end{tabular}
\caption{Comprehensive ablation results of GroundTeller on NuPrompt. We analyze the contribution of each core design component and the impact of different backbone architectures. UG: UniScene Generation; LAOD: Language-Aligned Object Decoding; TR: Temporal Reasoning; HR: Historical Reasoning; TGP: Temporal Grounding Prediction. \textbf{Bold} indicates the best performance.}
\label{tab:final_ablation}
\end{table*}

\subsection{Main Results}
Table~\ref{tab:full_metrics_mtml} and Fig.~\ref{fig:main results} report the quantitative comparison between TrackTeller and representative baselines under different filtering thresholds $\tau$. TrackTeller consistently outperforms all competing methods across evaluation metrics, demonstrating the effectiveness of jointly modeling multimodal perception and temporal reasoning for prompt-guided 3D tracking. As shown in Table~\ref{tab:full_metrics_mtml}, TrackTeller achieves the highest AMOTA and Recall at all thresholds, with particularly large gains under stricter filtering (e.g., $\tau=0.2$ and $0.3$), indicating stronger robustness to missing prompt-relevant targets. It also yields consistently lower AMOTP, confirming more accurate 3D localization.

In addition, TrackTeller significantly reduces identity-related errors, as reflected by lower TID and FAF across thresholds. Compared with PromptTrack and PromptTrack-3D, these improvements show that explicitly modeling short-term temporal context helps preserve object identity under occlusion, viewpoint changes, and weak visual evidence, while better suppressing prompt-irrelevant distractors. Fig.~\ref{fig:main results} further illustrates performance trends as $\tau$ increases. Although all methods degrade with fewer effective targets, TrackTeller exhibits a substantially slower drop in AMOTA and Recall. While PromptTrack-3D benefits from LiDAR inputs, it still lags behind TrackTeller, indicating that multimodal sensing alone is insufficient. The superior performance of TrackTeller stems from its unified design that tightly couples language alignment, temporal reasoning, and multimodal fusion.

Finally, we also compare model complexity and inference efficiency, with detailed results in the Appendix \ref{complex}. TrackTeller achieves a favorable balance of model size and real-time inference speed.

\begin{figure*}[t]
    \centering
    \includegraphics[width=1\linewidth]{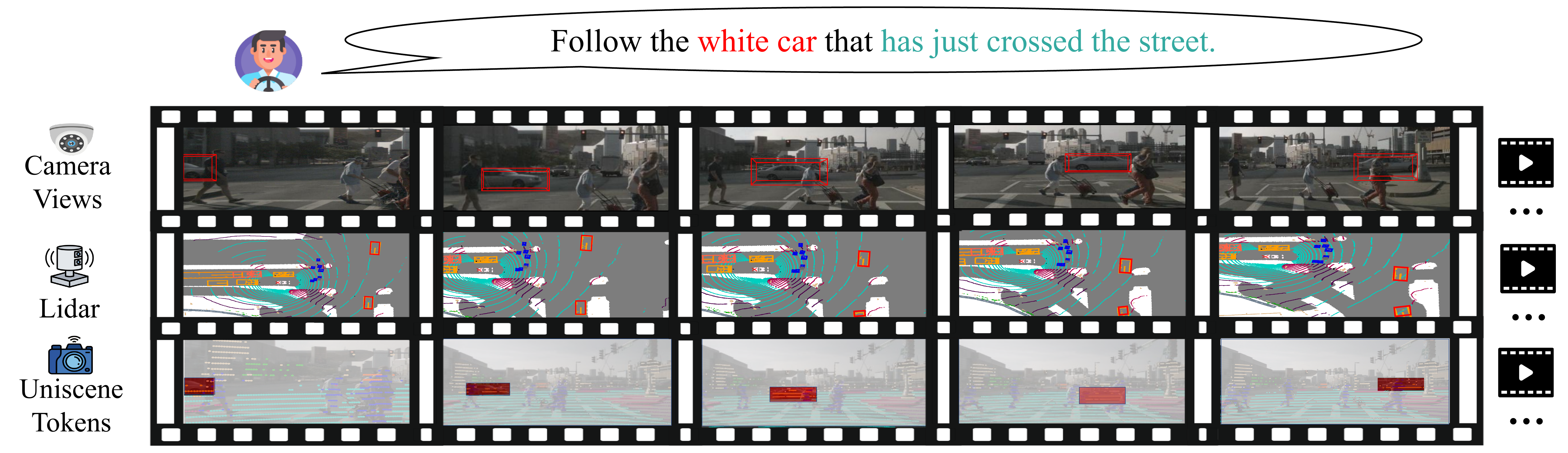}
    \caption{Qualitative comparison of grounding results using different modalities.}
    \label{fig:case}
\end{figure*}

\subsection{Ablation Studies}
\label{ablation}

We conduct ablation studies with $\tau=0.3$ to evaluate the contribution of key components in GroundTeller. As summarized in Table~\ref{tab:final_ablation}, introducing UniScene generation with LiDAR fusion consistently improves AMOTA and Recall over the image-only baseline, demonstrating the benefit of incorporating 3D geometric cues. Adding the language-aligned decoding modules (LSM and LGD) further yields substantial gains, particularly in Recall, highlighting the importance of explicit language-conditioned proposal generation. These quantitative gains are consistent with the qualitative analysis in the Appendix~\ref{vis}.


We progressively enable TR components. HR improves temporal stability, as indicated by lower TID, while TGP further increases AMOTA and Recall by preserving object continuity under weak observations. We also evaluate different backbones. Replacing VoVNet with ResNet-101 degrades performance, and CLIP text embeddings perform slightly worse than RoBERTa, suggesting that the gains mainly stem from the proposed architecture. Additional backbone comparisons are reported in Appendix~\ref{app:backbone}.

\subsection{Case Study}
Fig.~\ref{fig:case} presents a driving scenario with motion cues. Image-only grounding is affected by depth ambiguity and viewpoint changes, while LiDAR-only grounding lacks appearance information, resulting in confusion among nearby vehicles. TrackTeller fuses multimodal cues into UniScene tokens and applies temporal reasoning across frames, enabling consistent localization of the referred car based on both appearance and recent motion, even under weak or partial visual evidence.

\section{Conclusion}
This paper introduces TrackTeller, a prompt-guided temporal 3D grounding framework that jointly reasons over language, multimodal perception, and temporal dynamics.
By constructing UniScene tokens to integrate image and LiDAR cues, aligning language with scene representations, and incorporating historical and future-aware temporal reasoning, TrackTeller effectively resolves behavior-dependent referring expressions in dynamic driving scenarios.
Experiments demonstrate consistent improvements over strong baselines in tracking accuracy and temporal stability.


\bibliography{custom}

\appendix
\label{sec:appendix}

\section{Implementation of the $\mathrm{PreHead}$. }
\label{app:prehead}
$\mathrm{PreHead}(\cdot)$ maps the refined decoder queries into explicit 3D bounding box coordinates and their corresponding language-grounding scores. The design of $\mathrm{PreHead}(\cdot)$ follows the principles of query-based detection frameworks while being specifically optimized to handle language-conditioned feature embeddings. The input is $Z_L \in \mathbb{R}^{M \times D}$. Each query $z_i \in \mathbb{R}^D$ represents a localized feature vector that has integrated language-aware geometric evidence through the LGD iterative process. $\mathrm{PreHead}(\cdot)$ is implemented as a lightweight, parallelized structure consisting of two specialized branches: a \textit{3D Regression Branch} and a \textit{Grounding Classification Branch}. Both branches are built using MLPs.

\begin{itemize}
    \item \textbf{3D Regression Branch:} It transforms each query $z_i$ into a 3D bounding box $\hat{b}_i \in \mathbb{R}^k$. The box representation $\hat{b}_i$ includes the center coordinates $(x, y, z)$, dimensions $(w, l, h)$, and yaw orientation $\theta$. Specifically, the transformation is formulated as:
    \begin{equation}
        \hat{b}_i = \mathrm{MLP}_{\text{reg}}(z_i), \quad B_t^{\text{det}} = \{ \hat{b}_i \}_{i=1}^M.
    \end{equation}
    To ensure stable convergence, the branch predicts relative offsets with respect to the reference points associated with each query.

    \item \textbf{Grounding Classification Branch:} It estimates the semantic alignment between each query and the referring expression. This branch produces a scalar score $\hat{s}_i$ indicating the probability that the $i$-th proposal matches the target described in the text:
    \begin{equation}
        \hat{s}_i = \sigma(\mathrm{MLP}_{\text{cls}}(z_i)), \quad S_t^{\text{frame}} = \{ \hat{s}_i \}_{i=1}^M,
    \end{equation}
    These scores provide the initial frame-level grounding evidence used for subsequent temporal reasoning.
\end{itemize}


\section{Implementation of the $\mathrm{FutureReg}$.}
\label{appendix:future_reasoning}
$\mathrm{FutureReg}(\cdot)$ is designed to extrapolate object hypotheses from historical observations and provide temporally grounded proposals for the current frame. Unlike historical reasoning, which refines object representations using current visual evidence $U_t^{\text{lang}}$, future reasoning performs predictive inference purely based on past motion and temporal context. This predictive capability is essential for maintaining object continuity during periods of occlusion or when sensory evidence is weak.

Given the historical embeddings $H_t \in \mathbb{R}^{M \times D}$ that represent the object states over a temporal window, the module follows a two-stage process: temporal extrapolation and proposal regression. First, a temporal transformer $\mathcal{T}_{\text{fut}}(\cdot)$ is employed to model long-range dependencies within $H_t$. To distinguish between different time steps, we introduce learnable temporal position embeddings that encode the relative indices of historical frames. By treating the historical embeddings as keys and values while initializing a set of future queries corresponding to the target time step $t$, the transformer captures motion trends and temporal correlations via multi-head self-attention. This results in a set of future-aware embeddings: $F_t = \mathcal{T}_{\text{fut}}(H_t),$ where $F_t \in \mathbb{R}^{M \times D}$ represents the predicted latent state of the proposals for the current frame.

In the second stage, these future embeddings $F_t$ are processed by a lightweight regression head, $\mathrm{RegHead}(\cdot)$, which consists of multiple fully connected layers. This head decodes the motion-aware features into explicit geometric parameters and confidence cues, yielding the temporally propagated proposals and their associated scores:
\begin{equation}
B_{t}^{\text{his}},S_{t}^{\text{his}} = \mathrm{RegHead}(F_t).
\end{equation}
where $B_{t}^{\text{his}}$ corresponds to the predicted 3D bounding boxes in the current frame, while $S_{t}^{\text{his}}$ provides the confidence of these predictions based on the reliability of the temporal extrapolation. By operating independently of the current-frame visual features, the $\mathrm{FutureReg}(\cdot)$ module provides a robust set of hypotheses that complement the detection-based results from the LGD module, effectively stabilizing object trajectories across challenging driving scenarios.

\section{Data Examples}
\label{dataex}
\begin{figure}[htbp]
    \centering
    \begin{subfigure}[b]{0.4\textwidth}
        \centering
        \includegraphics[width=\textwidth]{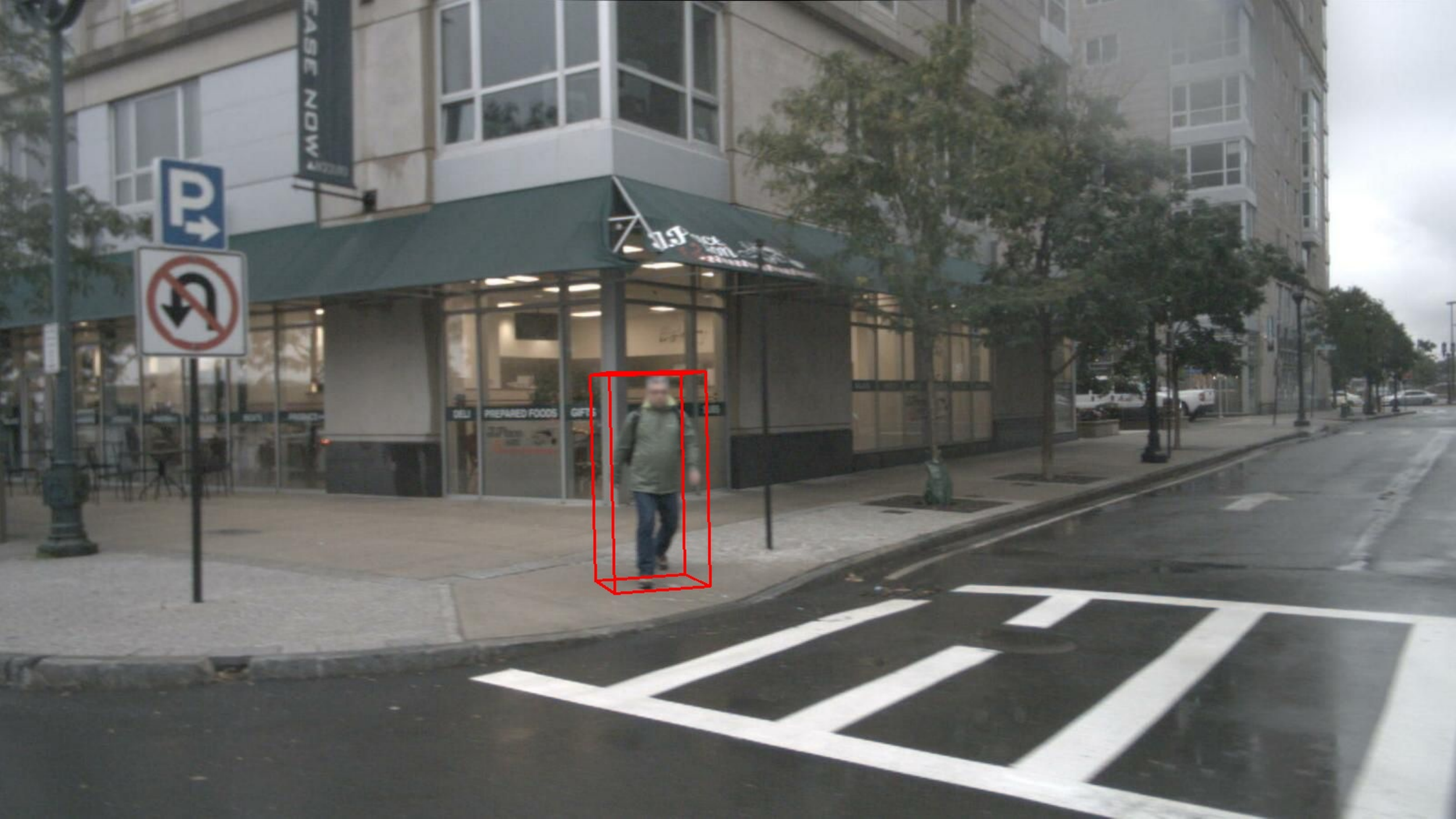} 
        \caption{Frame $t$ (FRONT RIGHT View)}
    \end{subfigure}
    \hfill
    \begin{subfigure}[b]{0.4\textwidth}
        \centering
        \includegraphics[width=\textwidth]{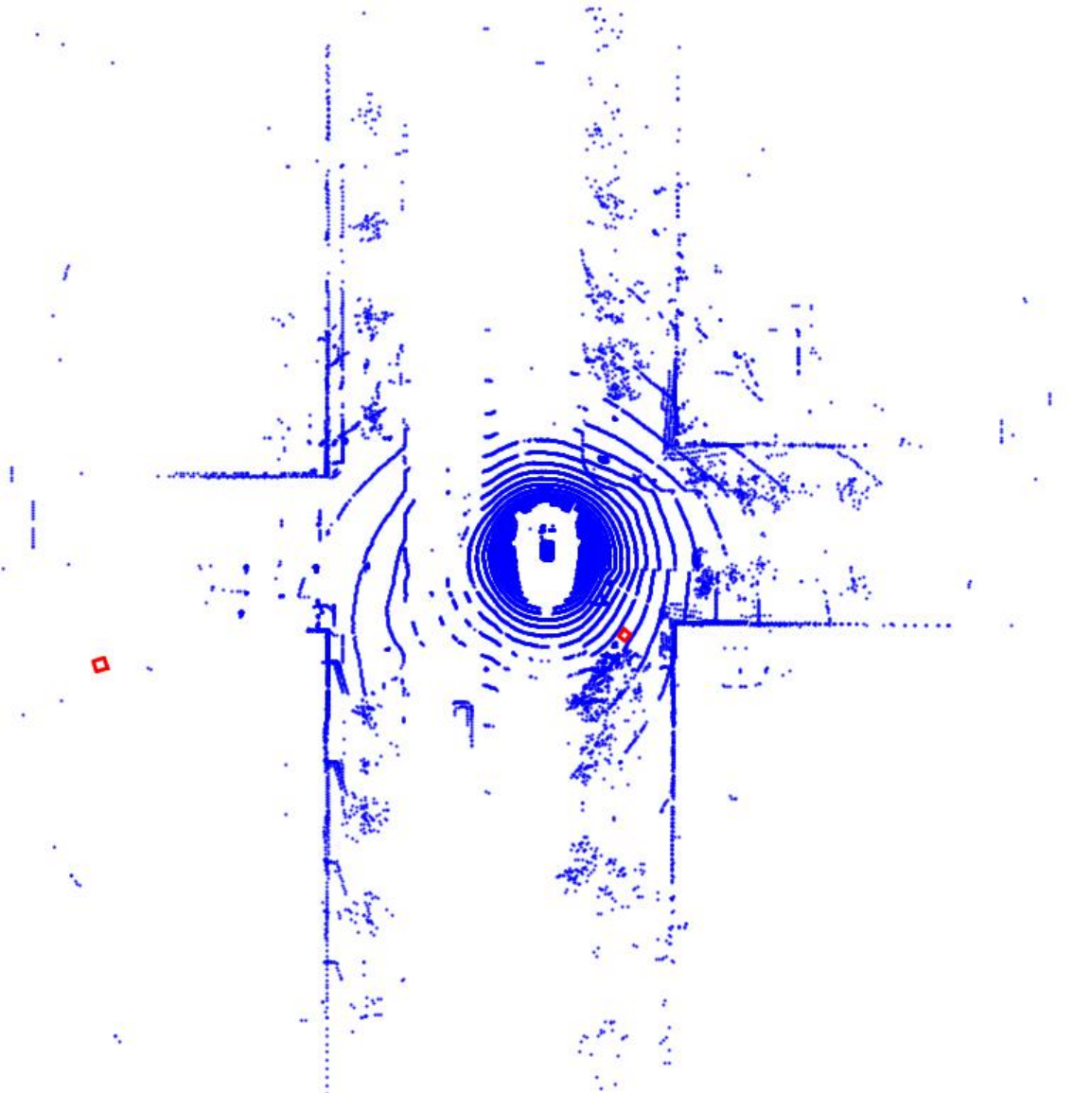} 
        \caption{Frame $t$ (BEV)}
    \end{subfigure}
    \hfill
    \begin{subfigure}[b]{0.4\textwidth}
        \centering
        \includegraphics[width=\textwidth]{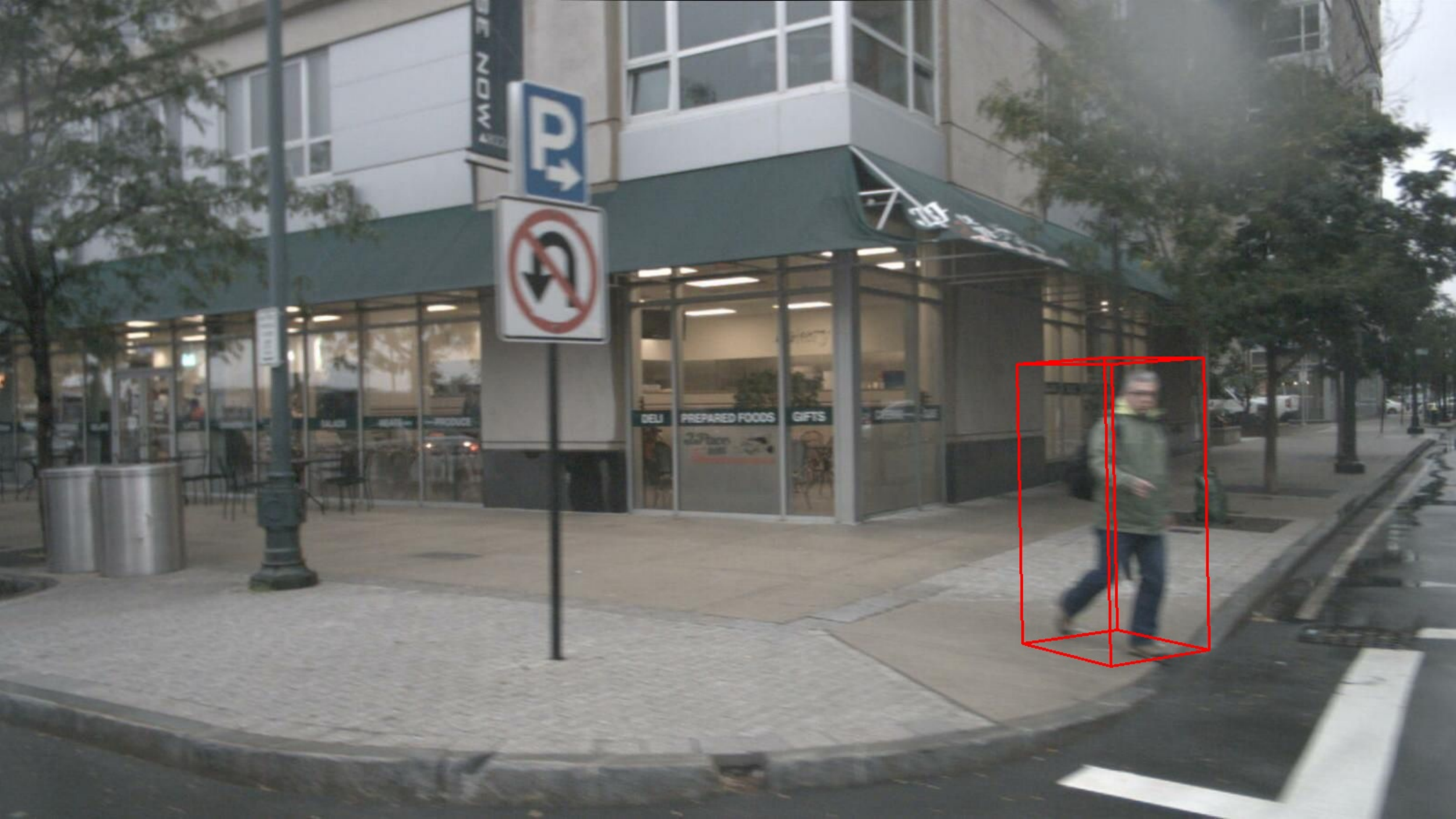}  
        \caption{Frame $t+1$ (FRONT RIGHT View)}
    \end{subfigure}
    \hfill
    \begin{subfigure}[b]{0.4\textwidth}
        \centering
        \includegraphics[width=\textwidth]{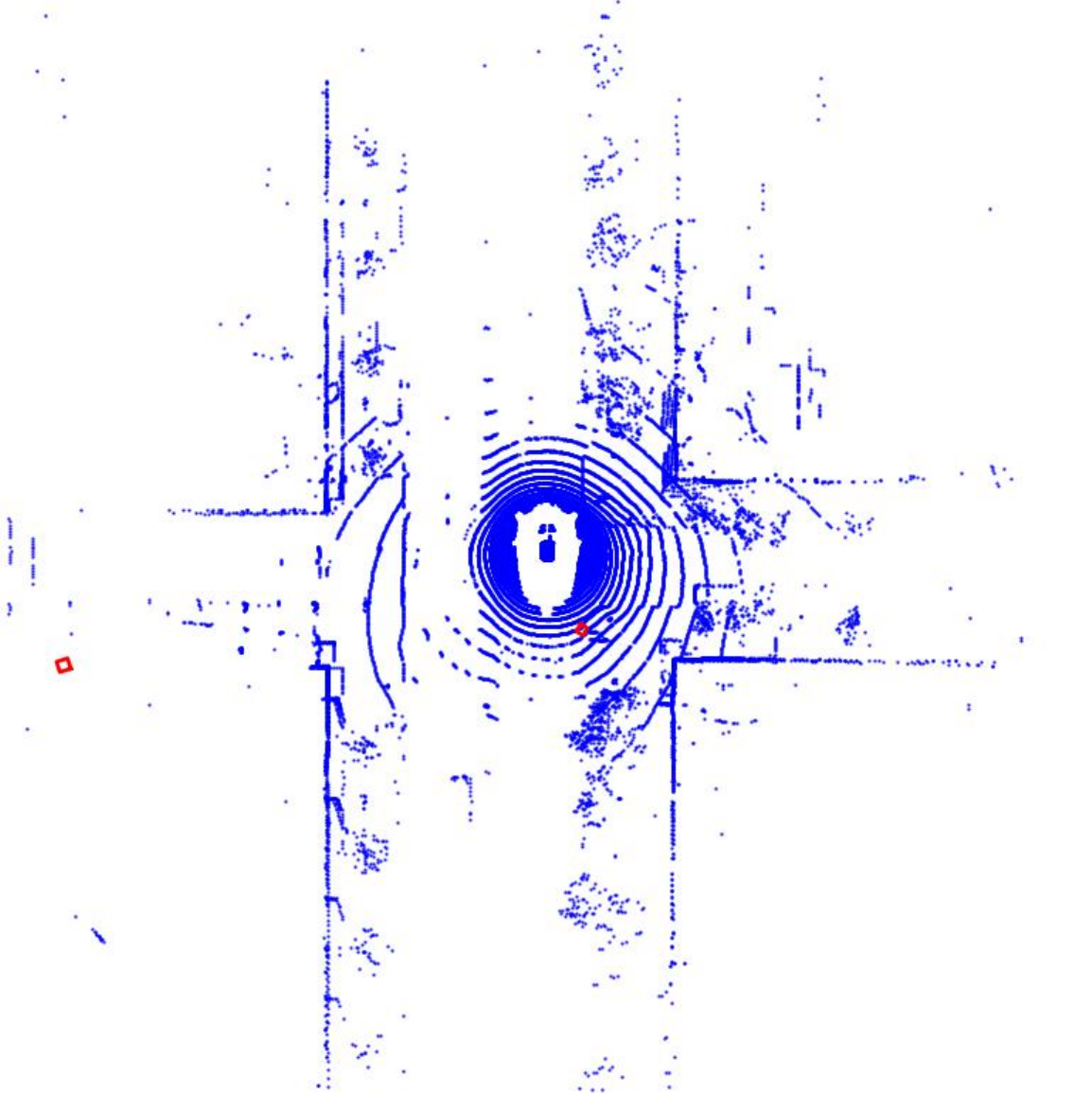}
        \caption{Frame $t+1$ (BEV)}
    \end{subfigure}

    \caption{Visualization of temporal sequences. The language prompt guides the model to focus on specific object tokens (e.g., ``pedestrian in blue pants'') across multiple frames, even as their relative positions change.}
    \label{fig:temporal_example}
\end{figure}

Fig.~\ref{fig:temporal_example} and Table~\ref{tab:data_structure} present an example of our multimodal temporal data representation. Each sequence comprises synchronized camera images, LiDAR-based BEV maps, and natural-language prompts that consistently refer to the target object over time, forming the basis for prompt-guided temporal grounding.

\begin{table}[htbp]
    \centering
    \small
    \setlength{\tabcolsep}{6pt}
    \renewcommand{\arraystretch}{1.15}

    \begin{tabular}{p{1.2cm}|p{5.6cm}}
        \toprule
        \textbf{Field} & \textbf{Content / Example} \\
        \midrule
        \textbf{Scene Token} &
        \texttt{0ac05652a4c44374998be876ba5cd6fd} \\
        \midrule
        \multirow{3}{*}{\textbf{Prompts}} &
        ``An individual in blue pants'' \\
        &
        ``The pedestrian wearing blue pants'' \\
        &
        ``The person walking with blue trousers'' \\
        \bottomrule
    \end{tabular}

    \caption{Structural breakdown of the language-conditioned temporal data, illustrating the correspondence between scene tokens and multiple semantically equivalent language prompts.}
    \label{tab:data_structure}
\end{table}


\section{Parameter Sensitivity Analysis}
\label{app:param_sensitivity}
We analyze the sensitivity of our model to several key hyperparameters. Rather than exhaustively evaluating all possible design choices, we focus on the most influential training-related parameters, including the learning rate, $\lambda_{\text{det}}$, and $\lambda_{\text{fut}}$. In each experiment, only one hyperparameter is varied, while all other settings are kept fixed according to the default configuration described in Section~\ref{sec:implementation details}.

\begin{table}[t!]
\centering
\small
\setlength{\tabcolsep}{1pt}
\begin{tabular}{c|c|ccccc}
\toprule
Parameter & Value & AMOTA$\uparrow$ & Rec.$\uparrow$ & AMOTP$\downarrow$ & TID$\downarrow$ & FAF$\downarrow$ \\
\midrule
\multirow{3}{*}{LR}
 & $1{\times}10^{-4}$ & 18.47 & 26.35 & 1.65 & 8.08 & 271.4 \\
 & $2{\times}10^{-4}$ & \textbf{18.79} & \textbf{28.42} & \textbf{1.63} & \textbf{8.07} & \textbf{231.4} \\
 & $4{\times}10^{-4}$ & 18.69 & 25.57 & 1.65 & 9.23 & 254.0 \\
\midrule
\multirow{3}{*}{$\lambda_{\text{det}}$}
 & 1.0 & 18.12 & 26.84 & 1.66 & 8.41 & 263.8 \\
 & 2.0 & \textbf{18.79} & \textbf{28.42} & \textbf{1.63} & \textbf{8.07} & \textbf{231.4} \\
 & 3.0 & 18.53 & 27.10 & 1.64 & 8.26 & 248.9 \\
\midrule
\multirow{3}{*}{$\lambda_{\text{fut}}$}
 & 0.25 & 18.41 & 27.65 & 1.64 & 8.32 & 245.6 \\
 & 0.5 & \textbf{18.79} & \textbf{28.42} & \textbf{1.63} & \textbf{8.07} & \textbf{231.4} \\
 & 1.0 & 18.36 & 26.98 & 1.65 & 8.89 & 259.7 \\
\bottomrule
\end{tabular}
\caption{Sensitivity analysis of key hyperparameters. LR refers to the learning rate.}
\label{tab:param_sensitivity}
\end{table}

As shown in Table~\ref{tab:param_sensitivity}, TrackTeller demonstrates stable performance under moderate variations of the selected hyperparameters. When varying the learning rate, the model achieves the best overall performance at $2\times10^{-4}$, while both smaller and larger learning rates result in slight degradation in tracking accuracy and temporal consistency. This suggests that the model is not highly sensitive to the learning rate within a reasonable range and reliably converges under the default setting.

In addition, we analyze the effect of $\lambda_{\text{det}}$. Reducing $\lambda_{\text{det}}$ weakens the supervision from object detection, leading to lower AMOTA and recall. On the other hand, excessively increasing $\lambda_{\text{det}}$ introduces a bias toward detection objectives, slightly degrading temporal metrics such as TID and FAF. The default value of $\lambda_{\text{det}} = 2.0$ strikes a favorable balance between detection accuracy and temporal tracking performance.

Finally, we examine the sensitivity to the $\lambda_{\text{fut}}$. A smaller value of $\lambda_{\text{fut}}$ limits the contribution of future supervision, resulting in inferior temporal consistency. Conversely, a larger value tends to over-regularize the model, negatively impacting both tracking accuracy and false alarm rates. The results indicate that TrackTeller is robust to variations in $\lambda_{\text{fut}}$, with $\lambda_{\text{fut}} = 0.5$ providing the best trade-off between future reasoning and current detection performance.

\section{Evaluation Metrics}
\label{EM}
Consistent with prior work on prompt-guided tracking, we use the Average Multi-Object Tracking Accuracy (AMOTA) as the primary evaluation metric. AMOTA captures overall tracking performance by averaging accuracy across a predefined set of recall thresholds and is defined as:
\begin{align}
\mathrm{AMOTA}
&= \frac{1}{|\mathcal{R}|}
   \sum_{r \in \mathcal{R}}
   \max \Bigl(
      0,\,
      1 - \notag \\
&\quad
      \frac{\mathrm{FN}_r + \mathrm{FP}_r + \mathrm{IDS}_r}
           {\mathrm{GT}_r}
   \Bigr),
\end{align}
where $\mathcal{R}$ denotes the set of recall thresholds, and $\mathrm{FN}_r$, $\mathrm{FP}_r$, $\mathrm{IDS}_r$, and $\mathrm{GT}_r$ correspond to the numbers of false negatives, false positives, identity switches, and ground-truth objects at threshold $r$, respectively. Given the language-conditioned nature of our task, which emphasizes prompt-aware per-object tracking rather than global multi-class identification, we focus on prompt-consistent identity preservation when interpreting identity-related errors.

To evaluate spatial precision, we report the Average Multi-Object Tracking Precision (AMOTP), which measures the mean 3D localization error of matched trajectories across recall thresholds:
\begin{equation}
\mathrm{AMOTP}
= \frac{1}{|\mathcal{R}|}
  \sum_{r \in \mathcal{R}}
  \frac{\sum_{i,t} d_{i,t}^r}{\mathrm{TP}_r},
\end{equation}
where $d_{i,t}^r$ denotes the Euclidean distance between the predicted 3D bounding box and the corresponding ground-truth box for a matched pair $(i,t)$ at threshold $r$, and $\mathrm{TP}_r$ is the number of true positive matches at threshold $r$.

In addition, we report Recall, which measures the proportion of prompt-referred ground-truth objects that are successfully tracked. To maintain consistency with AMOTA and AMOTP, Recall is computed by averaging over recall thresholds:
\begin{equation}
\mathrm{Recall}
= \frac{1}{|\mathcal{R}|}
  \sum_{r \in \mathcal{R}}
  \frac{\mathrm{TP}_r}{\mathrm{GT}_r}.
\end{equation}

To assess the model’s ability to maintain temporal identity consistency along a prompt-referred trajectory, we introduce the Temporal Identity Discontinuity (TID). It measures the average temporal gap between consecutive matched segments of a tracked object, thereby quantifying trajectory fragmentation:
\begin{equation}
\mathrm{TID}
= \frac{1}{N_{\text{matched}}}
  \sum_{i=1}^{N_{\text{matched}}}
  \sum_{k=1}^{K_i - 1}
  \Delta t_{i,k},
\end{equation}
where $N_{\text{matched}}$ denotes the number of prompt-referred ground-truth objects that are successfully matched, $K_i$ is the number of fragmented segments for the $i$-th object, and $\Delta t_{i,k}$ represents the duration of the $k$-th temporal gap between consecutive segments.

Finally, we report the False Alarm Frequency (FAF), which measures the average number of false positive predictions per frame and reflects the model’s ability to suppress prompt-irrelevant distractors:
\begin{equation}
\mathrm{FAF}
= \frac{1}{T}
  \sum_{t=1}^{T}
  \mathrm{FP}_t,
\end{equation}
where $T$ denotes the total number of frames in the sequence and $\mathrm{FP}_t$ is the number of false positives at frame $t$.

\section{Implementation of Baseline Models}
\label{app:baseline_heads}

Table~\ref{tab:baseline_comparison} summarizes the compared methods and their core design features.
\begin{table}[h]
\centering
\small
\setlength{\tabcolsep}{5.5pt} 
\begin{tabular}{l|l|l}
\hline
\textbf{Method} & \textbf{Key Feature} & \textbf{Modality} \\
\hline
DQTrack & Decoupled Queries  & Img  \\
PFTrack & Past\&Future Reasoning  & Img \\
PromptTrack & Prompt Reasoning & Img  \\
PromptTrack-3D & BEV Fusion  & LiD+Img  \\
\hline
\end{tabular}
\caption{Overview of representative baseline methods.}
\label{tab:baseline_comparison}
\end{table}

All baseline methods adopt two query-based tracking heads, namely DETR3D~\cite{detr3d} and PETR~\cite{liu2022petrv2}. DETR3D associates sparse 3D object queries with multi-view image features through camera geometry, while PETR encodes 3D positional information into image features to enable position-aware query attention for 3D detection. In both cases, prompt cross-attention is applied to refine head input, following the design of Prompt4Driving~\cite{wu2025language}.

\section{Complexity and Inference Efficiency}
\label{complex}
In addition to tracking accuracy, we further compare the model complexity and inference efficiency of the evaluated methods.
Table~\ref{tab:model_complexity} reports the number of parameters and the inference speed measured in frames per second (FPS).
Camera-only methods generally exhibit lower parameter counts due to the absence of point cloud encoders, while multimodal approaches introduce additional overhead from voxel-based or point-based LiDAR backbones.
Despite integrating both image and LiDAR modalities, TrackTeller maintains a competitive model size and achieves real-time inference speed, benefiting from efficient cross-modal fusion and prompt-guided decoding.
These results demonstrate that the proposed method strikes a favorable balance between tracking performance and computational efficiency.

\begin{table}[t]
\centering
\scriptsize
\setlength{\tabcolsep}{6pt}
\begin{tabular}{l|l|c|c}
\toprule
\textbf{Modality} & \textbf{Method} & \textbf{Params (M)$\downarrow$} & \textbf{FPS$\uparrow$} \\
\midrule
Img
& DQTrack (DETR3D)      & 820 & 3.9 \\
Img
& DQTrack (PETR)        & 894 & 3.8 \\
\midrule
Img
& PF-Track (DETR3D)     & 883 & 3.8 \\
Img
& PF-Track (PETR)       & 961 & 3.7 \\
\midrule
Img
& PromptTrack (DETR3D)  & 1413 & 2.8 \\
Img
& PromptTrack (PETR)    & 1485 & 2.8 \\
\midrule
Img+LiD
& PromptTrack-3D (DETR3D) & 1559 & 2.7 \\
Img+LiD
& PromptTrack-3D (PETR)   & 1631 & 2.7 \\
\midrule
Img+LiD
& \textbf{TrackTeller (Ours)} & 1632 & 2.7 \\
\bottomrule
\end{tabular}
\caption{Comparison of model complexity and inference efficiency.
Params denote the number of learnable parameters, and Frame Per Second (FPS) is measured under the same hardware configuration.}
\label{tab:model_complexity}
\end{table}

\section{Qualitative Analysis of Language-Aligned Object Decoding}
\label{vis}
\begin{figure}[t]
    \centering
    \begin{subfigure}[b]{0.36\textwidth}
        \centering
        \includegraphics[width=\textwidth]{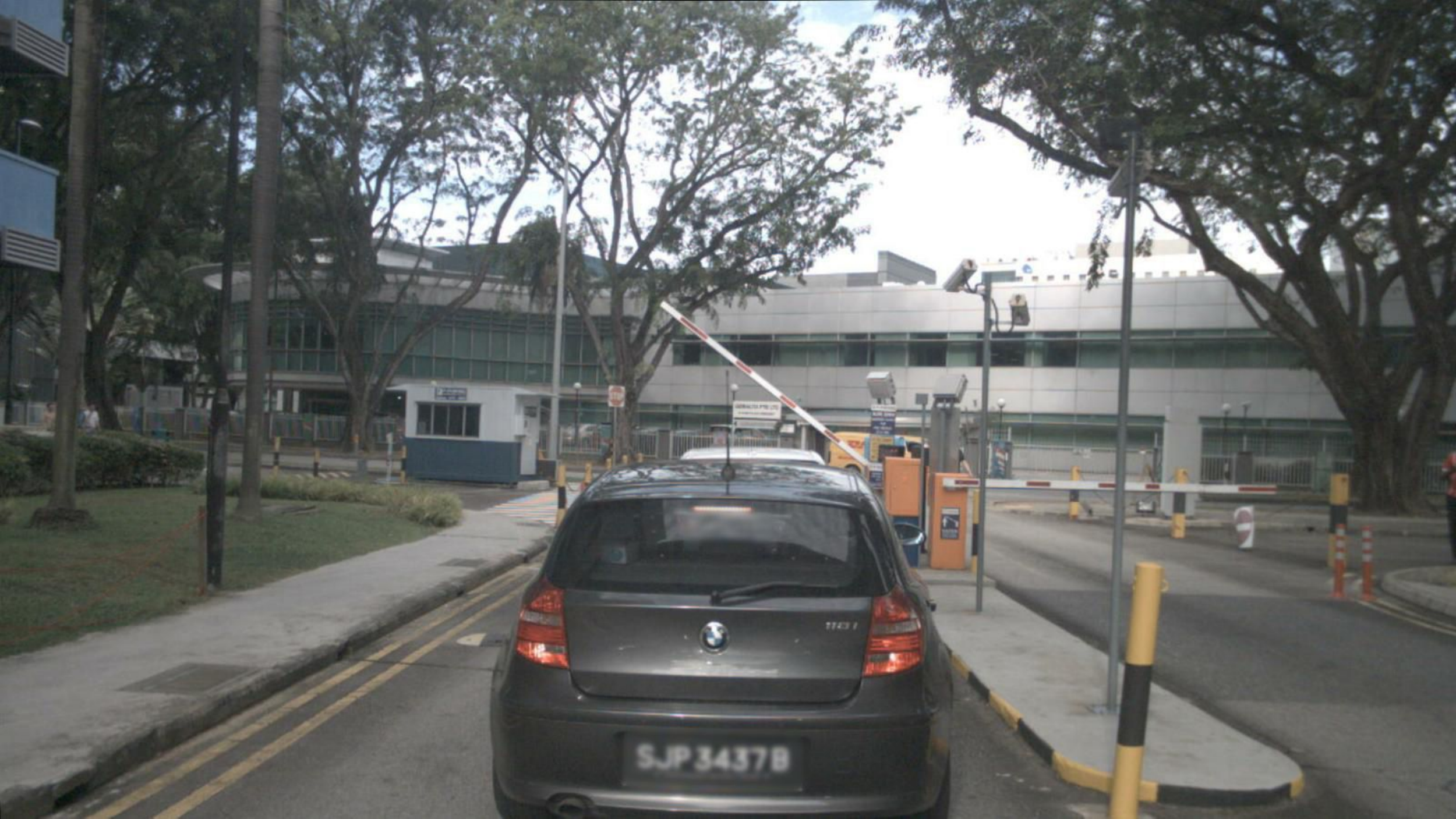}
        \caption{Original image}
        \label{fig:original_image}
    \end{subfigure}
    \hfill
    \begin{subfigure}[b]{0.36\textwidth}
        \centering
        \includegraphics[width=\textwidth]{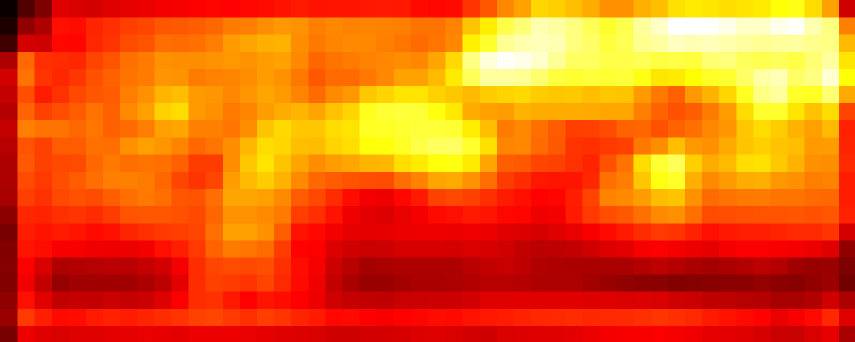}
        \caption{Output of the LSM}
        \label{fig:lsm_feature}
    \end{subfigure}
    \hfill
    \begin{subfigure}[b]{0.36\textwidth}
        \centering
        \includegraphics[width=\textwidth]{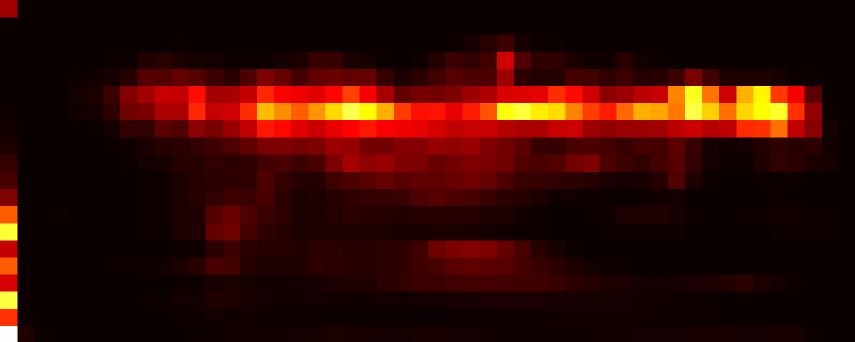}
        \caption{Output of the LGD}
        \label{fig:lgd_feature}
    \end{subfigure}
    \caption{
    Visualization of the original image and the feature maps produced by LSM and LGD, illustrating how the model progressively refines attention from general semantic alignment to prompt-specific target localization.}
    \label{fig:feature_visualization}
\end{figure}

To better understand how the proposed modules operate on UniScene tokens, we visualize feature representations at different stages of the model (Fig. \ref{fig:feature_visualization}). Specifically, we present the original image, the feature map produced by the LSM module, and the feature map produced by the LGD module. In this example, the prompt instructs the model to ``follow the black car ahead.''

\begin{table*}[htbp]
\centering
\small
\setlength{\tabcolsep}{6pt}
\caption{Performance comparison across different encoder backbones for various modalities.}
\label{tab:ec}
\begin{tabular}{c|c|ccccc}
\toprule
Image Encoder & Prompt Encoder &
AMOTA$\uparrow$ & Rec.$\uparrow$ & AMOTP$\downarrow$ & TID$\downarrow$ & FAF$\downarrow$ \\
\midrule
VoVNet & RoBERTa & \textbf{50} & \textbf{35} & \textbf{10} & \textbf{10} & \textbf{160} \\
\midrule
VoVNet & BERT & 17.86 & 27.31 & 1.72 & 8.44 & 182.9 \\
VoVNet & DeBERTa & 18.05 & 27.68 & 1.71 & 8.29 & 176.4 \\
VoVNet & DistilBERT & 17.42 & 26.54 & 1.73 & 8.71 & 195.8 \\
VoVNet & CLIP (Text) & 15.42 & 25.20 & 1.74 & 8.59 & 221.3 \\
\midrule
ResNet-101 & RoBERTa & 13.15 & 21.80 & 1.75 & 9.12 & 265.4 \\
Swin Transformer & RoBERTa & 7.86 & 12.66 & 1.86 & 12.69 & 345.7 \\
\bottomrule
\end{tabular}
\end{table*}

The LSM feature map exhibits broadly distributed attention across the scene, indicating that the model is primarily performing coarse semantic alignment without precise target localization. In contrast, the LGD output shows a clear concentration of attention on the prompt-referred object. The model effectively associates the linguistic description with both appearance and contextual cues, resulting in focused activation on the black car. These visualizations demonstrate how LSM and LGD progressively refine the representation from general language-scene alignment to prompt-specific target localization.

\section{Effect of Different Encoder Choices}
\label{app:backbone}
We investigate the impact of different encoder choices for image, point cloud, and language modalities. All experiments use the same configuration as described in Section~\ref{sec:implementation details}, with only the encoder being replaced while keeping all other components unchanged. This approach allows us to isolate the effect of different encoder architectures on performance across multiple modalities.

Table \ref{tab:ec} demonstrates the effectiveness of the proposed framework in achieving high performance across all metrics. The combination of VoVNet as the image encoder and RoBERTa as the prompt encoder provides the best overall results, excelling in AMOTA, Recall, and TID. This combination highlights the robustness of our framework in integrating different modalities effectively, delivering stable and accurate performance. Although variations in the encoder choice lead to some differences in performance, our framework remains strong even with alternatives like BERT and DeBERTa, which also show competitive results. In contrast, using a Swin Transformer with RoBERTa leads to lower performance, particularly in recall and tracking accuracy. Additionally, the use of the CLIP encoder, while less effective, still showcases the flexibility of our framework to accommodate different encoder types. These findings further support the versatility of our approach, which can adapt to various encoder configurations without compromising core performance.

\end{document}